\documentclass{amia}
\usepackage{lipsum} 
\usepackage{csquotes}
\usepackage{multirow}
\usepackage{xcolor}
\usepackage{wrapfig}
\usepackage[superscript,biblabel]{cite}
\usepackage{makecell}
\usepackage{amssymb}
\usepackage{hyperref}
\usepackage{subfig}
\usepackage{comment}

\begin{document}

\title{Deep Representations of First-person Pronouns for Prediction of Depression Symptom Severity}

\author{Xinyang Ren, BSc$^1$, Hannah A Burkhardt, PhD$^1$,  Patricia A Areán, PhD$^2$, \\ Thomas D Hull, PhD$^3$,
Trevor Cohen, MBChB, PhD$^1$ }

\institutes{
    $^1$ Department of Biomedical Informatics and Medical Education, University of Washington, Seattle, WA; $^2$ ALACRITY Center, Department of Psychiatry and Behavioral Sciences, University of Washington, Seattle, WA; $^3$ Talkspace, New York, NY}

\maketitle

\section*{Abstract}

\textit{
Prior work has shown that analyzing the use of first-person singular pronouns can provide insight into individuals’ mental status, especially depression symptom severity. These findings were generated by counting frequencies of first-person singular pronouns in text data. However, counting doesn’t capture how these pronouns are used. Recent advances in neural language modeling have leveraged methods generating contextual embeddings. In this study, we sought to utilize the embeddings of first-person pronouns obtained from contextualized language representation models to capture ways these pronouns are used, to analyze mental status. De-identified text messages sent during online psychotherapy with weekly assessment of depression severity were used for evaluation. Results indicate the advantage of contextualized first-person pronoun embeddings over standard classification token embeddings and frequency-based pronoun analysis results in predicting depression symptom severity. This suggests contextual representations of first-person pronouns can enhance the predictive utility of language used by people with depression symptoms.}

\section*{Introduction}
More than 50 million Americans experienced a mental illness between 2019 and 2020 \cite{reinert2021state}. Furthermore, coronavirus disease (COVID-19)-related experiences during the pandemic increased the prevalence of anxiety and depression \cite{gallagher2020impact}. Depression is also the most prevalent cause of disability globally\cite{liu2020changes}. Thus, mental health services are in great demand. To address the longstanding shortage of mental health professionals\cite{thomas2009county}, particularly in areas outside major urban centers, there is a desire to leverage technology to provide more patients with access to therapy, with increasing acceptance of digital delivery platforms in the wake of the pandemic\cite{ben2020digital}. In particular, initial evidence suggests that text messaging may be an effective modality for the delivery of psychotherapeutic interventions for depression \cite{senanayake2019effectiveness}. The rapid adoption of text-based therapy presents unprecedented opportunities for linguistic analysis, with the potential to support the development of methods that provide insight into the trajectories of patients' mental states that may serve as markers or even predictors of treatment efficacy or failure. 
Prior work has shown the difference in language use between depressed and never-depressed individuals, especially with respect to the frequency of first-person singular pronoun use. Depressed individuals are thought to have higher self-focused attention \cite{ingram1984depression, pyszczynski1987self}, and have been shown to use more first-person singular pronouns (e.g. ``I", ``me", ``my", ``myself") in written essays than never-depressed individuals \cite{rude2004language}. This difference has since been observed across a variety of studies\cite{holtzman2017meta}. 

These findings emerged from automated analysis of language from study participants. To analyze the mental status of an individual based on what he or she said or wrote, the predominant method involves calculating the percentage of words belonging to pre-defined psychologically meaningful categories. Linguistic Inquiry and Word Count (LIWC) \cite{tausczik2010psychological} is a commonly used text analysis software program that applies this method to quantify linguistic indicators of mental status in text. A considerable amount of prior work has utilized LIWC to analyze depression in social media posts\cite{coppersmith-etal-2014-quantifying} and online therapy chat text\cite{burkhardt2021behavioral}, amongst other sources \cite{holtzman2017meta}. LIWC's function word categories include first-person pronouns (e.g. ``I", ``me", ``my", ``myself"). While previous research findings have indicated a positive correlation between depression and first-person singular pronoun usage measured by LIWC\cite{holtzman2017meta}, a recent study analyzing language used by healthcare workers under stress on Reddit during the peak of the COVID-19 pandemic has revealed a \textit{decreasing} trend in first-person singular pronoun use, which the authors argue reflects a way of self-distancing from traumatic events \cite{ireland2022tracking}. This suggests that the ways in which first-person pronouns are used may provide information beyond that which can be obtained by simply counting their frequencies.

Contextualized language representation models such as Bidirectional Encoder Representations from Transformers (BERT)\cite{DBLP} have reached state-of-the-art performance on many downstream natural language processing (NLP) tasks\cite{9271752}. These models produce embeddings of each word (or subword) token using an encoder with a self-attention mechanism to add information from context words to the  representation of a given token.
The encoder network has been pre-trained on tasks such as masked language modeling and next sentence prediction\cite{DBLP} with large amounts of data. Prior work has found that these models can provide contextually modulated representations that reflect distinct linguistic categories \cite{petersen2023lexical}. This too suggests that the contextual embeddings of first-person pronouns (e.g. ``I", ``me", ``my", ``myself") may capture information about the way in which these pronouns are used that is of value for analysis of  mental status. 
  
To evaluate this hypothesis, we explored how well the contextual embeddings of first-person pronouns (``I", ``me", ``my", ``myself", ``mine") obtained from MentalBERT\cite{ji2021mentalbert}, a BERT-based language model further pre-trained for the domain of mental healthcare, represent mental status. Specifically, we used the average of the contextual embeddings of the first-person pronouns from patient-authored text to predict depression symptom severity. We applied our method to online therapy chat logs accompanied by 9-item Patient Health Questionnaire (PHQ-9) scores \cite{kroenke2001phq}. The PHQ-9 is a validated, self-administrated questionnaire measuring depression severity with scores ranging from 0 to 27, providing a validated measure of depression symptom severity with established diagnostic thresholds\cite{moriarty2015screening}. We evaluated the utility of the resulting representations as predictive features for PHQ-derived depression status, in comparison with both LIWC-derived features and representations derived from BERT without restricting its focus to personal pronouns. 

\section*{Methods}
\label{sec_methods}
\textbf{Data} 

De-identified text messages sent during message-based psychotherapy collected in an ongoing study\cite{arean2021protocol} were utilized in our work. Participants communicated with therapists using a secure messaging platform provided by Talkspace, a digital mental health company, over 12 weeks. All participants consented to the use of their transcript and study-related data for research purposes. The participants completed weekly PHQ-9 questionnaires\cite{kroenke2001phq}. The study received ethics approval from the University of Washington Institutional Review Board.

In the current work, only messages sent by clients were used. We aggregated the messages within one week preceding PHQ-9 administration, or since the last score if two PHQ-9 scores were administered within the same one-week period. We discarded aggregated messages shorter than 30 tokens. Tokenization is the process of breaking textual data into discrete elements called tokens. With BERT's tokenizer, these tokens may be individual words. However, less frequently occurring words may be subdivided into smaller units. The number of tokens is obtained from the BERT model tokenizer. As the maximum length limit of tokens for the BERT model is 512, we truncated those messages longer than 510 tokens (to accommodate the addition of special tokens such as [CLS] during preprocessing) into several shorter messages with a maximum token length of 300. For example, if the message token length was 800, we truncated it into 3 messages with token length 300, 300, and 200, and assigned the same PHQ-9 score to each of these messages. We excluded participants who had fewer than 4 PHQ-9 scores available. After preprocessing, we had data from 94 out of 201 participants. The demographic information and average PHQ-9 score in each demographic category of these 94 participants is provided in Table \ref{demographic}. On average, each participant had 6.21 PHQ-9 scores available and wrote 685.67 words weekly. 

Participants also completed daily surveys consisting of six Ecological Momentary Assessment (EMA) questions related to depression-related symptoms and participation in pleasurable activities (``When did you fall asleep last night?", ``When did you wake up today?", ``How difficult was it to fall and stay asleep last night?", ``What best describes your activity level today?", ``Did you do anything social today?", ``How much did you enjoy yourself?"). We collected EMA responses within the same time range as the message data (one week preceding PHQ-9 administration) and calculated the median scores for each time range. 

\newcommand*{\MyIndent}{\hspace*{0.5cm}}%
\begin{table}[ht]
\small
\begin{center}
\begin{tabular}{|l c c|} 
 \hline
Demographic Categories & Counts  & Average PHQ-9 score\\ [0.5ex] 
 \hline
 Age &  &\\ 
 \MyIndent 18-22 & 27 (28.7$\%$) & 10.94 \\
 \MyIndent 23-27 & 21 (22.3$\%$) & 10.83\\
 \MyIndent 28-32 & 15 (16$\%$)& 7.76\\
 \MyIndent 33-37 & 11 (11.7$\%$) & 9.96\\
 \MyIndent 38+   & 20 (21.3$\%$) & 12.04\\
 Gender & &\\
 \MyIndent Female & 78 (83$\%$) & 10.45\\
 \MyIndent Male   & 13 (13.8$\%$) & 10.8\\ 
 \MyIndent Transgender Male & 1 (1.1$\%$) & 12.04\\
 \MyIndent Something else  & 2 (2.1$\%$) & 8.73\\
 Race & &\\
 \MyIndent White or Caucasian & 62 (65$\%$) & 10.44\\
 \MyIndent Black or African American & 9 (9.6$\%$) & 8.89\\
 \MyIndent Asian   & 8 (8.5$\%$) & 11.95\\
 \MyIndent American Indian/Alaskan Native + Caucasian & 2 (2.1$\%$) & 7.21\\
 \MyIndent Asian + Caucasian & 2 (2.1$\%$) & 11.68\\
 \MyIndent Prefer to self-describe & 7 (7.4$\%$) & 11.13\\
 \MyIndent Prefer not to say & 4 (4.3$\%$) & 12.7\\
Primary Language & &\\
 \MyIndent English & 91 (96.8$\%$) & 10.53\\
 \MyIndent Other (Arabic,Tagalog,Spanish) & 3 (3.2$\%$) & 10.52\\
 \hline
\end{tabular}
\caption{Demographic information and average PHQ-9 score of the 94 participants}
\label{demographic}
\end{center}
\end{table}

\textbf{Algorithm design}

\begin{figure}[ht]
  \centering
    \includegraphics[width=0.7\textwidth]{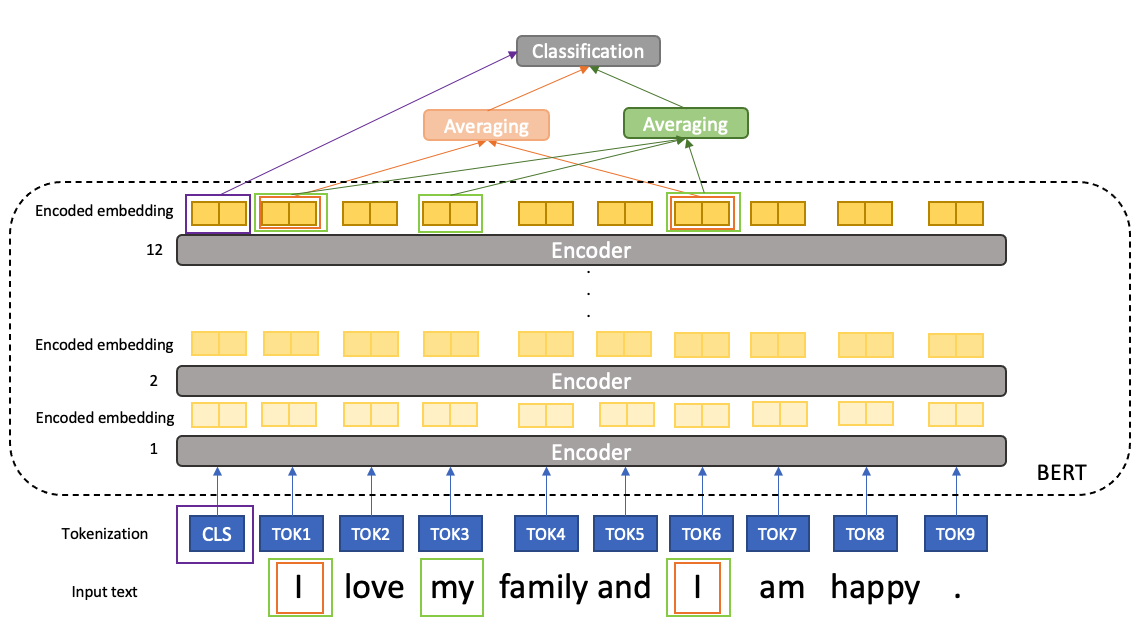}
 \caption{Workflow of the baseline model (purple squares and arrows) using the embedding of the [CLS] token, pronoun classifier model that uses the average of ``I" embeddings (orange squares and arrows), and pronoun classifier model that uses the average of ``I", ``me", ``my", ``myself", ``mine" embeddings (green squares and arrows).}
 \label{fig:algorithm}
\end{figure}

The goal of our study was to evaluate the utility of contextual representations of first-person pronouns (``I", ``me", ``my", ``myself", ``mine") as predictors of depression symptom severity. In addition, we evaluated the utility of representations of the pronoun ``I" in isolation, on account of the seminal finding that frequency of use of this pronoun increases with depression severity \cite{rude2004language}. First, we evaluated whether contextual representations of personal pronouns captured information pertinent to prediction of depression status beyond that provided by the standard BERT aggregation process of using the contextualized representation of a specialized classification token ([CLS]) prepended to a sequence.
The baseline model is therefore the BertForSequenceClassification model\cite{bert}, which appends a binary classification head (in the form of a fully-connected layer) to a pretrained BERT model. Our BERT pronoun classifier model extends this baseline model by forcing it to focus on personal pronouns at the point of classification. Instead of providing BERT’s output representation of the classification token ([CLS]) to the classification head, we use the average of all the personal pronoun token embeddings. We use contextual embeddings from BERT's last layer because prior work found that BERT captures semantic features at higher layers\cite{jawahar2019does}. For example, when considering the personal pronoun ``I" in isolation, we take the average of the ``I" token embeddings from BERT’s last layer as the output for the classification task. For input data that do not contain ``I" or ``i", we manually insert a single ``I" at the beginning of the text. To capture the way in which first-person pronouns are used fully, we designed an extended version of our BERT pronoun classifier model that uses the average of the ``I", ``me", ``my", ``myself", ``mine" token embeddings from the BERT's last layer as the output for the classification task. Input data that do not contain ``I" or ``i" were processed in the same way (prepending a single ``I") for fair comparison between the two pronoun classifier models. The workflow of the baseline model and our pronoun classifier models is shown in Figure \ref{fig:algorithm} with example input text.

The pre-trained BERT model utilized is MentalBERT \footnote{available in the HuggingFace repository as "mental/mental-bert-base-uncased"}, which was developed to improve performance on mental health related NLP tasks \cite{ji2021mentalbert}. MentalBERT is based on the bert-based-uncased model\cite{DBLP}, which consists of 12 transformer layers with 110M parameters in total. Starting from this base model (which has already been trained on text from the internet and digitized books), MentalBERT was further pre-trained on mental health related Reddit posts. In our experiments, we focus on the contextual embeddings provided by this pre-trained model without it being further fine-tuned for classification (though we do train the weights of the classification head, which were not involved in the pre-training process, and are randomly initialized ahead of classification training). As the number of parameters in language representation models is large - 110M in our case - the fine-tuning process is  computationally expensive. Moreover, fine-tuning this large number of parameters may obscure differences in the contributions of the original contextual embeddings to the depression symptom severity prediction as the model is updated for task-specificity. However, as we aimed to characterize the relationship between fine-tuning and the quality of pronoun-focused representations, we also conducted secondary experiments in which we did not freeze the pre-trained layers of the model.

\textbf{Experimental setup}

As PHQ-9 scores were collected weekly, messages were aggregated in the one-week period preceding PHQ-9 score collection (as described previously). We used messages aggregated with the most recent PHQ-9 score as the test data set (n=200). We used the messages with the first three PHQ-9 scores available to construct the training and validation data sets (n=870). We randomly split these messages into 5 folds, and ran each of our experiments 5 times. In each of these five runs, one fold was reserved as the validation data set and the remaining 4 folds formed the training data set. The prediction goal is a binary outcome derived form the PHQ-9 total score. A prior meta-analysis study found that at cut-off score 10, PHQ-9 score can achieve acceptable diagnostic performance in different clinical settings\cite{moriarty2015screening}. Therefore, we regarded a PHQ-9 score larger than or equal to 10 (positive class) as indicative of major depressive disorder (MDD) and a PHQ-9 score of less than 10 (negative class) as indicating that a participant was not depressed. The class distribution in the training and validation data sets is n$_{pos}$=461 and n$_{neg}$=409. The class distribution in the test data sets is n$_{pos}$=109 and n$_{neg}$=91. To explore how the pronoun classifier models' performance differs between textual data that contain first-person pronouns and those that do not, we further constructed a data set consisting only of messages that contain ``I" or ``i" using the same training, validation, and test split setting.

We used the \texttt{HuggingFace} implementation of BERT \cite{wolf2020transformers} \footnote{source code at HuggingFace repository "BertForSequenceClassification"\cite{bert}}. To train the model, we used a learning rate of 1x10\textsuperscript{-5} with a 0.1 warm-up proportion. We used the Adam optimizer\cite{kingma2014adam} and trained the model for a maximum of 10 epochs, utilizing an early stopping strategy such that if model performance on the validation data set (evaluated by macro-averaged F1 score) no longer improved after 4 continuous epochs of training, the training was stopped. We saved models while performance on the validation data set improved during training, and then evaluated the performance of the saved best model on the test data set. The experiments were conducted using one NVIDIA v100 GPU.

The LIWC analysis results of all messages were obtained by running the LIWC-22 software program\cite{boyd2022development}. A logistic regression classifier from the scikit-learn package\cite{pedregosa2011scikit} was applied to the entire spectrum of LIWC analysis variables including features with known correlation with depression symptom severity, such as the frequencies of positive and negative emotion words as defined in LIWC's lexicon\cite{boyd2022development}, as well as the LIWC ``i" variable (first-person singular pronouns such as ``I", ``me", ``my", ``myself") as an isolated feature. We used the same train-test split and the five experimental run settings as the BERT models. We pre-processed the data with scaling (center to the mean and scale to unit variance). The optimization method utilized was the limited-memory BFGS method\cite{liu1989limited} with L2 regularization.
\section*{Results}
\textbf{Without fine-tuning} 

The performance of the baseline model and our proposed BERT pronoun classifier models \textit{without} the pre-trained component of the BERT model being fine-tuned on the downstream task is shown in Table \ref{without_ft}. The table provides the average macro-averaged F1 score, F1 score of the positive class, accuracy, Area Under the Receiver Operator Characteristic Curve (AUROC), and Area Under the Precision Recall Curve (AUPRC) on the test data set for each of the 5 experimental runs. The results indicate that the contextual embeddings of first-person pronouns carry information for prediction of depression symptom severity that exceeds that provided by the classification token ([CLS]) embedding. 
\begin{table}[ht]
\small
\begin{center}
\begin{tabular}{|p{25mm} | p{40mm}| c|c | c | c| c|} 
\hline
Data & Embeddings  & F1$_{macro}$	& F1$_{positive}$ & Accuracy	& AUROC &  AUPRC\\
\hline
\multirow{3}{25mm}{Messages that contain ``I" or ``i"} & CLS & 0.503 & 0.637 & 0.549  & 0.542 & 0.587 \\
 & ``I" & 0.572\textsuperscript{\textdagger} &  0.662 &  0.592\textsuperscript{\textdagger} & 0.612\textsuperscript{\textdagger} & 0.638\textsuperscript{\textdagger}\\
 & ``I",``me",``my",``myself",``mine" & \textbf{0.586}\textsuperscript{\textdagger} & \textbf{0.681} &\textbf{0.612}\textsuperscript{\textdagger} & \textbf{0.624}\textsuperscript{\textdagger} & \textbf{0.648}\textsuperscript{\textdagger} \\
 \hline
\multirow{3}{25mm}{All messages} & CLS & 0.513 & 0.619 & 0.538  & 0.509 & 0.543 \\
 & ``I" & \textbf{0.584}\textsuperscript{\textdagger} &  0.653 &\textbf{0.604}\textsuperscript{\textdagger} & \textbf{0.62}\textsuperscript{\textdagger} & \textbf{0.647}\textsuperscript{\textdagger} \\
 & ``I",``me",``my",``myself",``mine" & 0.575\textsuperscript{\textdagger}& \textbf{0.672} & 0.6\textsuperscript{\textdagger} & 0.638\textsuperscript{\textdagger} & 0.658\textsuperscript{\textdagger}\\
 \hline
\end{tabular}
\end{center}
\caption{Performance of the baseline model ([CLS]), pronoun classifier model that uses the average of ``I" embeddings, and pronoun classifier model that uses the average of ``I", ``me", ``my", ``myself", ``mine" embeddings without the pre-trained BERT component of the model being fine-tuned on the test data set. $\dagger$ indicates statistical significance in difference between the average of proposed model and baseline model five-run performance by paired t-test.}
\label{without_ft}
\end{table}

\textbf{With fine-tuning}

To investigate how fine-tuning affects model performance, we fine-tuned the pre-trained BERT component of the three models using the same experimental settings. The results are shown in Table \ref{with_ft}, which can be compared with Table \ref{without_ft}. The performance of the baseline model has been considerably improved by fine-tuning. While the BERT pronoun classifier models also show improved performance, they have not benefited as much. As we anticipated, enabling fine-tuning has obscured the difference between the models. Though the pronoun-focused models do perform better across the majority of configurations and metrics when models are fine-tuned, these improvements over the baseline models are no longer statistically significant across the five repeated runs. Of note, we also found no statistical significance in the difference between the \textit{fine-tuned} baseline model's performance and the \textit{sans-fine-tuning} BERT pronoun classifier models' performance using a paired t-test. This indicates that the advantages in performance conferred by focusing on personal pronoun embeddings while training the classifier head only is comparable to that conferred by fine-tuning the BERT component of the baseline model while training the classifier. This is important on account of the computational advantages of avoiding fine-tuning BERT end-to-end. In our experiments, even with a relatively small version of BERT, this fine-tuning increased training time approximately threefold. 
\begin{table}[ht]
\small
\begin{center}
\begin{tabular}{|p{25mm} | p{40mm}| c |c| c | c| c|} 
\hline
Data & Embeddings & F1$_{macro}$	& F1$_{positive}$ & Accuracy	& AUROC &  AUPRC\\
\hline
\multirow{3}{25mm}{Messages that contain ``I" or ``i"} & CLS & 0.586 & 0.617 & 0.591  & 0.628  & 0.692 \\
 & ``I" & \textbf{0.594} & 0.646 & \textbf{0.604} & \textbf{0.653} & \textbf{0.72} \\
 & ``I",``me",``my",``myself",``mine" & 0.592 & \textbf{0.647} & 0.6 & 0.647 & 0.712 \\
 \hline
\multirow{3}{25mm}{All messages} & CLS & \textbf{0.589} & 0.624  &  \textbf{0.591} & 0.636  & 0.686 \\
 & ``I" & 0.581 & \textbf{0.64} & 0.588 & \textbf{0.678} & 0.695 \\
 & ``I",``me",``my",``myself",``mine" & 0.577 & 0.615 & 0.582 & 0.635 & \textbf{0.701} \\
 \hline
\end{tabular}
\end{center}
\caption{Average performance (n=5) of the baseline model ([CLS]), pronoun classifier model that uses the average of ``I" embeddings, and pronoun classifier model that uses the average of ``I", ``me", ``my", ``myself", ``mine" embeddings with fine-tuning on the test data set.}
\label{with_ft}
\end{table}

\textbf{Baseline: LIWC analytics}

 The logistic regression classifier performance using LIWC analysis features are shown in Table \ref{lr_liwc}. In contrast to prior work using messages from text-based therapy\cite{burkhardt2022comparing}, LIWC variables appear to offer limited utility as predictors of PHQ-9 derived depression status in this set.

\begin{table}[ht]
\small
\begin{center}
\begin{tabular}{|c | p{3cm} | c | c | c| c|c|} 
\hline
LIWC Analysis & Model & F1$_{macro}$ & F1$_{positive}$ & Accuracy	& AUROC &  AUPRC\\
\hline
 LIWC all & \multirow{2}{3cm}{Logistic regression classifier} & 0.477 & 0.515 & 0.48 & 0.465 & 0.54\\
 \cline{3-7}
 LIWC ``i" &  & 0.382 & 0.688 & 0.534 & 0.4 & 0.471\\
 \hline
\end{tabular}
\end{center}
\caption{Performance of logistic regression classifier on LIWC analysis results of all message data on the test data set. "LIWC all" refers to the entire spectrum of LIWC analysis variables.}
\label{lr_liwc}
\end{table}

To further compare the LIWC ``i" variable and predictions from the contextual embeddings of first-person pronouns as indicators of mental status, we calculated their correlation with participants' EMA responses. We utilized the responses to the last four of these questions as they are ordinal in nature. The probabilities of the positive class predicted by the models trained on all messages for the PHQ-9 binary outcome classification task in the \textit{without fine-tuning} configuration for validation and test data sets were used as the BERT pronoun classifier model's predictions. A LIWC ``i" variable analysis of the same message data was obtained. The calculated correlation measures between the EMA responses and pronoun analysis results are shown in Table \ref{correlation}. The table also includes the results from paired t-tests on the pronoun analysis results smaller than and larger than or equal to the median response score (2, 1, 0.5, and 2 for the questions, respectively) to assess the statistical significance of the mean difference between the two groups, shown in the Mean$_{low}$ and Mean$_{high}$ columns of Table \ref{correlation}.

The predictions from the BERT pronoun models are generally more strongly and significantly correlated with EMA values than those from the LIWC pronoun variable, though neither model produces significant correlations (or differences between means when dichotomized) with the EMA question concerning daily activity levels. The strongest correlation is with the question concerning sleep difficulties, which are known to coexist with depression\cite{staner2010comorbidity}. An exception to this general rule concerns the last question, which solicits enjoyment. In this case, the LIWC ``i" variable shows a weak but significant positive correlation with enjoyment, which is consistent with prior literature showing associations between this variable and aspects of behavioral activation \cite{burkhardt2021behavioral}.

\begin{table}[ht]
\small
\begin{center}
\begin{tabular}{| p{3.3cm} | c | p{3.6cm}|p{2.4cm}| p{2.3cm}|} 
\hline
Question & Response Category & Pronoun Analysis & Mean$_{low}$, Mean$_{high}$ & Correlation measure using Kendall's $\tau$ \cite{kendall1938new} \\
\hline
\multirow{3}{*}{\makecell[l]{How difficult was it to \\fall and stay asleep \\last night? \\
(n=936, $\overline{n}_{weekly}$=2.5)}} & 
\multirow{3}{*}{\makecell[c]{Not at all: 0 \\ A little bit: 1 \\Somewhat: 2 \\Quite a bit: 3\\Very much: 4}}& LIWC ``i" variable & \makecell[c]{(8.919, 9.222)} &\makecell[c]{0.0329}\\
\cline{3-5}
 & & Predicted probability by BERT I model & \makecell[c]{(0.495, 0.505)$^{\bigstar}$}&\makecell[c]{0.1998$^*$}\\
 \cline{3-5}
 & & Predicted probability by BERT first-person model & \makecell[c]{(0.501, 0.514)$^{\bigstar}$} &\makecell[c]{0.2035$^*$} \\
 \hline
 \multirow{3}{*}{\makecell[l]{What best describes \\ your activity level \\ today?\\
 (n=723, $\overline{n}_{weekly}$=1.5)}} & 
\multirow{3}{*}{\makecell[c]{Sedentary: 0 \\Moderately active \\ for 30+ minutes: 1 \\Vigorously active\\ for 30+ minutes: 2}}& LIWC ``i" variable &\makecell[c]{(9.03, 8.931)} &\makecell[c]{-0.0223}\\
\cline{3-5}
 & & Predicted probability by BERT I model & \makecell[c]{(0.501, 0.499)}&\makecell[c]{0.0006}\\
 \cline{3-5}
 & & Predicted probability by BERT first-person model & \makecell[c]{(0.507, 0.507)}&\makecell[c]{0.0075}\\
 \hline
 \multirow{3}{*}{\makecell[l]{Did you do anything  \\ social today?\\
 (n=723, $\overline{n}_{weekly}$=1.5)}} & 
\multirow{3}{*}{\makecell[c]{No: 0 \\Yes: 1}}& LIWC ``i" variable & \makecell[c]{(9.127, 8.886)}&\makecell[c]{-0.0074}\\
\cline{3-5}
 & & Predicted probability by BERT I model & \makecell[c]{(0.506, 0.496)$^{\bigstar}$}&\makecell[c]{-0.0716$^\diamond$}\\
 \cline{3-5}
 & & Predicted probability by BERT first-person model & \makecell[c]{(0.512, 0.503)$^{\bigstar}$}&\makecell[c]{-0.1028$^\diamond$} \\
 \hline
 \multirow{3}{*}{\makecell[l]{How much did you\\enjoy yourself?\\
 (n=449, $\overline{n}_{weekly}$=1.3)}} & 
\multirow{3}{*}{\makecell[c]{Not at all: 0 \\ A little bit: 1 \\Somewhat: 2 \\Quite a bit: 3\\Very much: 4}}& LIWC ``i" variable & \makecell[c]{(8.268, 9.15)$^{\bigstar}$} &\makecell[c]{0.0769$^*$}\\
\cline{3-5}
 & & Predicted probability by BERT I model & \makecell[c]{(0.498, 0.496)}&\makecell[c]{-0.0453}\\
 \cline{3-5}
 & & Predicted probability by BERT first-person model &\makecell[c]{(0.505, 0.504)} &\makecell[c]{-0.0645$^\diamond$} \\
 \hline
\end{tabular}
\end{center}
\caption{Correlation between EMA responses and pronoun analysis results. In the Question column, n is the total number of question responses used for analysis and $\overline{n}_{weekly}$ is the average number of question responses within the one week time range. BERT I model refers to the BERT pronoun classifier model that uses average ``I" embeddings for classification. BERT first-person model refers to the BERT pronoun classifier model that uses the average of ``I", ``me", ``my", ``myself", ``mine" embeddings for classification. Mean$_{low}$ and Mean$_{high}$ refer to the mean of the pronoun analysis results smaller than and larger or equal to the median response score (2, 1, 0.5, and 2 for the questions respectively). $\bigstar$ indicates that the difference between 2 means is statistically significant (p-value $<$ 0.05). The correlation measures for the BERT I model and the BERT first-person model are the average value across the 5 experimental runs. $*$ indicates statistical significance on all runs (p-value $<$ 0.05). $\diamond$ indicates statistical significance in some but not all runs. }
\label{correlation}
\end{table}

\begin{figure}[ht]
  \centering
    \subfloat[][\centering]{\includegraphics[width=0.3\textwidth]{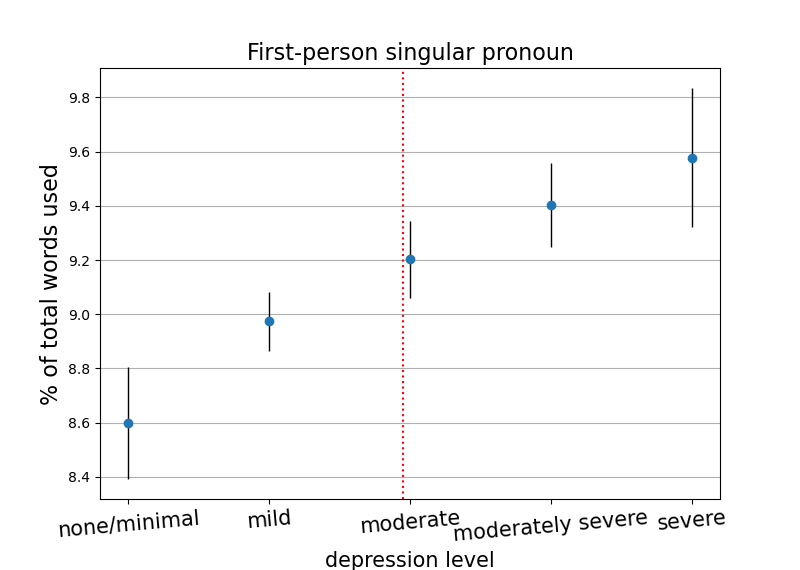}}%
    \qquad
    \subfloat[][\centering]{\includegraphics[width=0.3\textwidth]{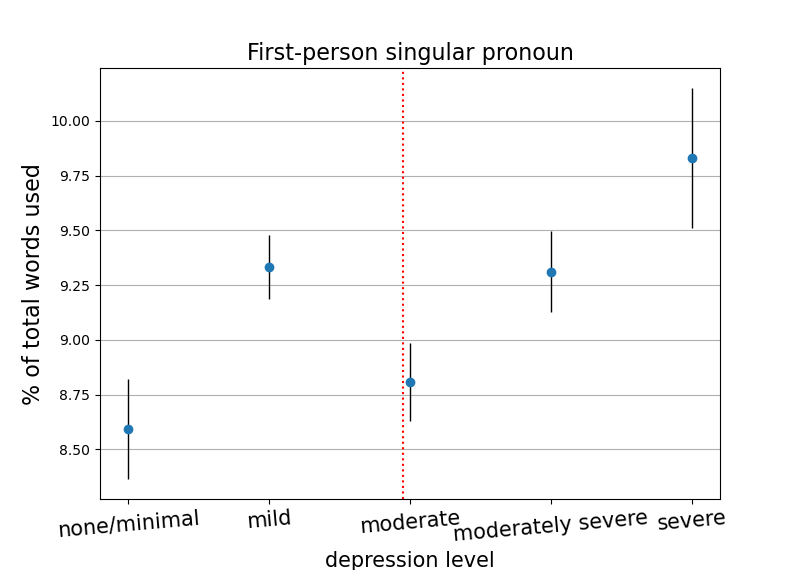}}%
    \qquad
    \subfloat[][\centering]{\includegraphics[width=0.3\textwidth]{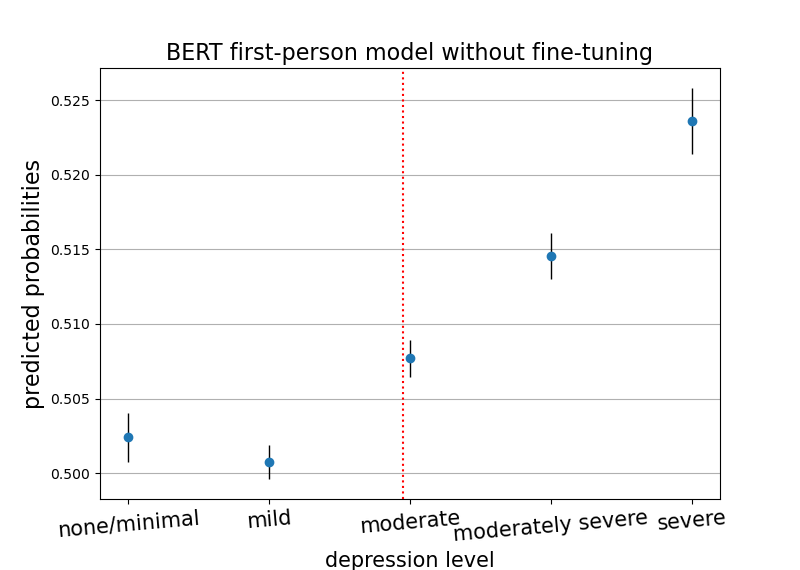}}%
 \caption{(a) Average of percentage of first-person singular pronouns used among total words in aggregated weekly messages with all PHQ-9 score records of the 94 participants. (b) Average of percentage of first-person singular pronouns used among total words in aggregated weekly messages with PHQ-9 score records in training and test data sets. (c) Average of the probabilities of positive class predicted by BERT first-person model with PHQ-9 score records in validation and test datasets in each depression level category\cite{spitzer2014test}: none/minimal (PHQ-9 $\in$ [0,5)), mild (PHQ-9 $\in$ [5,10)), moderate (PHQ-9 $\in$ [10,15)), moderately severe (PHQ-9 $\in$ [15,20)), severe (PHQ-9 $\in$ [20,27)). The error bar indicates the standard error of the mean. The red dotted vertical line indicates the cut-off score 10.}
 \label{fig:liwc_i}
\end{figure}

Figure \ref{fig:liwc_i} shows the average LIWC ``i" variable - the percentage of first-person singular pronouns among total words used - in aggregated weekly messages with all PHQ-9 score records of the 94 participants, with PHQ-9 score records in training and test data sets, and the average probabilities of the positive class predicted by the BERT pronoun classifier model using the average of ``I", ``me", ``my", ``myself" and ``mine" embeddings for classification in the \textit{without fine-tuning} configuration with PHQ-9 score records in validation and test data sets in each depression level category (the graph from the BERT pronoun classifier model using averaged ``I" embeddings for classification looks similar to plot (c) in Figure \ref{fig:liwc_i}, and is not included on account of space constraints). As shown in plot (a), participants with more severe depression symptoms tend to use more first-person singular pronouns, which is consistent with previous research findings\cite{holtzman2017meta}. At the cut-off score 10, plot (b) shows a decrease in the average LIWC ``i" variable while plot (c) shows the most rapid increase in the average predicted probabilities from the BERT pronoun classifier. This may explain the poor performance of the LIWC ``i" variable in the PHQ-9 binary outcome prediction task.

\section*{Discussion}
In this work, we evaluated the hypothesis that the contextual embeddings of first-person pronouns (e.g.``I", ``me", ``my", ``myself", ``mine") obtained from contextualized language representation models such as BERT\cite{DBLP} can provide information about first-person pronoun usage that is beneficial for mental status analysis. We designed two BERT pronoun classifier models, one using the average of ``I" embeddings and the other using the average of ``I", ``me", ``my", ``myself", and ``mine" embeddings for classification of depression status. Our results show the advantage of contextualized first-person pronoun embeddings over both standard classification token embedding and  frequency-based LIWC pronoun variables for this task. These findings indicate that pronoun embeddings encode contextual information about \textit{how} first-person pronouns are used that adds information beyond their usage frequencies that is indicative of mental status. This finding is in a sense not surprising on account of the many different contexts in which first-person pronouns may be used in text-based therapy sessions. Table \ref{quotes} provides some illustrative examples of use of the personal pronoun ``I" by participants with different PHQ-9 scores. The examples from participants with minimal symptoms (low PHQ-9 scores) describe positive sentiment and enjoyable activities. In contrast, those from participants above PHQ-9 score thresholds considered indicative of depression\cite{moriarty2015screening} (e.g. $>$= 10) show perceived lack of control, diminished trust, and anxiety. Of note, the \textit{frequency} of the use of this pronoun is highest in the example from the participant with the lowest PHQ-9 score, which is inconsistent with the strong evidence base for increasing use of first-person pronouns as depression severity increases\cite{holtzman2017meta}. In summary, these examples illustrate differences in contextual use of first-person pronoun that carry information pertinent to the assessment of mental state beyond that provided by pronoun frequency.

\begin{table}[ht]
\small
\begin{center}
\begin{tabular}{|c p{100mm}|} 
 \hline
Participant’s average PHQ-9 score & Quotes from messages  \\ [0.5ex] 
 \hline
 4.8 & \textit{\blockquote{I can honestly say I never felt better then I did yesterday. I did my daily survey last night and everything for the most part was positive.}} \\ 
 \hline
 8.5 & \textit{\blockquote{I usually journal or do something active like walking, playing basketball, running. I tried the 4-4-8 breath today as I laid outside after playing basketball. It was actually quite nice.}} \\
 \hline
 14.1 & \textit{\blockquote{i do feel like i have no control of any of this situation}} \\
 \hline
 18.3 & \textit{\blockquote{This is why I don’t trust anyone because whenever I need help no one is there for me. So now I rebuild mywalls and push everyone away.}} \\
 \hline
 23.3 & \textit{\blockquote{It’s made me fearful of things that didn’t scare me before and I get a lot of anxious intrusive thoughts as well as physical symptoms like irregular heartbeat and shaking}} \\  
 \hline
\end{tabular}
\end{center}
\caption{Quotes of messages written by participants with different average PHQ-9 score.}
\label{quotes}
\end{table}

Another key finding is the observation that the BERT pronoun classifier models can reach performance comparable to that of a fine-tuned BERT model using the standard ([CLS]) token for classification \textit{without} its BERT component being fine-tuned. This finding suggests that a priori selection of tokens that are likely to be informative for a task at hand may provide a general strategy through which to improve the computational efficiency of transformer-based text categorization that suggests a way to utilize large language models in the context of limited computational resources. This is particularly important with the advent of publicly released models that are large by current standards, such as BLOOM\cite{scao2022bloom}, and OPT\cite{zhang2022opt}. Training such models end-to-end is likely to be beyond the computational resources available in many research institutions. However, inference to generate embeddings may be feasible within available GPU or even CPU resources. By avoiding backpropagation through these large models, our approach of judicious embedding selection can provide an alternative with much lower computational cost. Moreover, recent studies\cite{misra2019black, lehman2021does} have found that it is possible to recover personal information in training data from the language models, posing great privacy concern in sharing such trained models. As many published language models like MentalBERT\cite{ji2021mentalbert} and BioBERT\cite{lee2020biobert} were pretrained on data that is publicly available, sharing models trained on the contextual embeddings obtained from these models without fine-tuning poses no additional privacy risks. However, the risk of leakage of personal information is a potential concern when models fine-tuned on tasks using sensitive data are shared. 

There are several limitations to this work. The mental health status of participants is evaluated solely on the basis of the self-reported PHQ-9 score, and we have excluded many participants due to data incompleteness. Future studies could take other measures such as the General Anxiety Disorder-7 score (GAD-7)\cite{williams2014gad} into account and utilize larger data sets to confirm our findings. Beyond the PHQ-9 binary outcome classification task, the contextual embeddings of first-person pronouns and the LIWC ``i" variable could be further compared in a regression analysis using the PHQ-9 total score that includes patient-specific variables to model longitudinal effects. While this will likely obscure differences due to diverging representations, we anticipate improvements in accuracy from models that take repeated measures into account. Finally, while we chose to use the contextual embeddings from BERT's last layer in this study, the representational utility of contextual pronoun embeddings obtained from other layers remains to be determined.   

\section*{Conclusion}
In this work we devised a richer representational alternative to LIWC's widely used frequency counting method to support analyses of first-person singular pronoun use in text data to infer mental health status, using contextual representations from the transformer deep learning architecture for natural language processing. The contextual embeddings of first-person singular pronouns resulted in better performance than the results of LIWC analysis as features for the prediction of depression symptom severity. In addition, our pronoun classification models achieved comparable performance to standard transformer-based categorization approaches, while avoiding computationally intensive end-to-end fine-tuning of large contextualized language representation models. This work suggests that contextualized representations of first-person pronouns capture information that is of value for linguistically informed models to infer mental state, with the potential to support diagnosis and treatment monitoring in the context of text-based therapy.

\subparagraph{Acknowledgments}
This research was supported by U.S. National Institute of Mental Health grants R44MH124334 and P50MH115837.
\makeatletter
\renewcommand{\@biblabel}[1]{\hfill #1.}
\makeatother

\bibliographystyle{vancouver}
\bibliography{amia}

\end{document}